\documentclass{article}

\usepackage{arxiv}

\usepackage[utf8]{inputenc} 
\usepackage[T1]{fontenc}    
\usepackage{hyperref}       
\usepackage{url}            
\usepackage{booktabs}       
\usepackage{amsfonts}       
\usepackage{nicefrac}       
\usepackage{microtype}      
\usepackage{graphicx}
\usepackage{algorithm}
\usepackage{algpseudocode} 
\usepackage{amsmath}     
\usepackage{amssymb}     
\usepackage{comment}
\usepackage{tabularx}
\usepackage{float}
\usepackage[flushleft]{threeparttable}

\newcommand{\dataset}{\mathcal{D}}
\newcommand{\batch}{\mathcal{B}}
\newcommand{\experts}{\mathcal{E}}
\newcommand{\params}{\boldsymbol{\theta}}

\title{EOE: Evolutionary Optimization of Experts \\ for Training Language Models}

\author{
  Yingshi Chen \\
  \texttt{gsp.cys@gmail.com} \\
}

\begin{document}
\maketitle 
\begin{abstract}
This paper presents an evolutionary framework for the training of large language models(LLM). The models are divided into several experts(sub-networks), which have the same structure but different parameter values. Only one expert is trained at each step. After the classical AdamW optimization, some evolutionary operators(crossover, PSO, and mutation) act on the tensor weights between the current expert and the best expert. So current expert would learn the experience of best expert. The direction of best expert would help current expert's loss decrease faster. Finally, only save the weight of the best expert.  Experiments show that best expert would achieve nearly the same accuracy as the full model. This would greatly reduce the size of the model for inference. Since only one expert is trained at each step, the training needs much less memory and has much higher throughput. Experiments show that the throughput would accelerate more than ten times!  Our source code is available at \url{https://github.com/gruai/koifish}. It's a pure c++/cu framework, which is suitable for easy deployment on PCs and edge computing devices. 

\end{abstract}

\keywords{Evolutionary computation \and Large language models \and Sparse LLM \and Crossover Mutation \and  Koifish }

\section{Introduction}
Large language models(LLM)\cite{vaswani2017attention, radford2019language, jiang2024mixtral,naveed2025comprehensive,comanici2025gemini,dai2024deepseekmoe} opens the door to new era. But it's too large and requires too many training resources. Even training a small model(~1B parameters) requires multiple GPUs. For example, it needs eight H100 GPUs and 24 hours to train a GPT-2 with 1558M parameters\cite{gpt2_example}. This poses a heavy burden on researchers and also hinders the rapid iteration and improvement of algorithms. A promising way to reduce train cost is using sparse structure, especially Mixture of Experts (MoE) architectures\cite{jiang2024mixtral,comanici2025gemini,dai2024deepseekmoe}. MoE is the cornerstone of Mixtral\cite{jiang2024mixtral}, Gemini\cite{comanici2025gemini}, DeepSeek\cite{dai2024deepseekmoe}, and more ... Each MoE layer has a gating network and multiple sub-networks (“experts”). When a token is processed, the gating network routes it to one or a few experts, and their outputs are combined as the final output. 

\begin{figure} [H]
    \centering
    \includegraphics[width=0.8\linewidth]{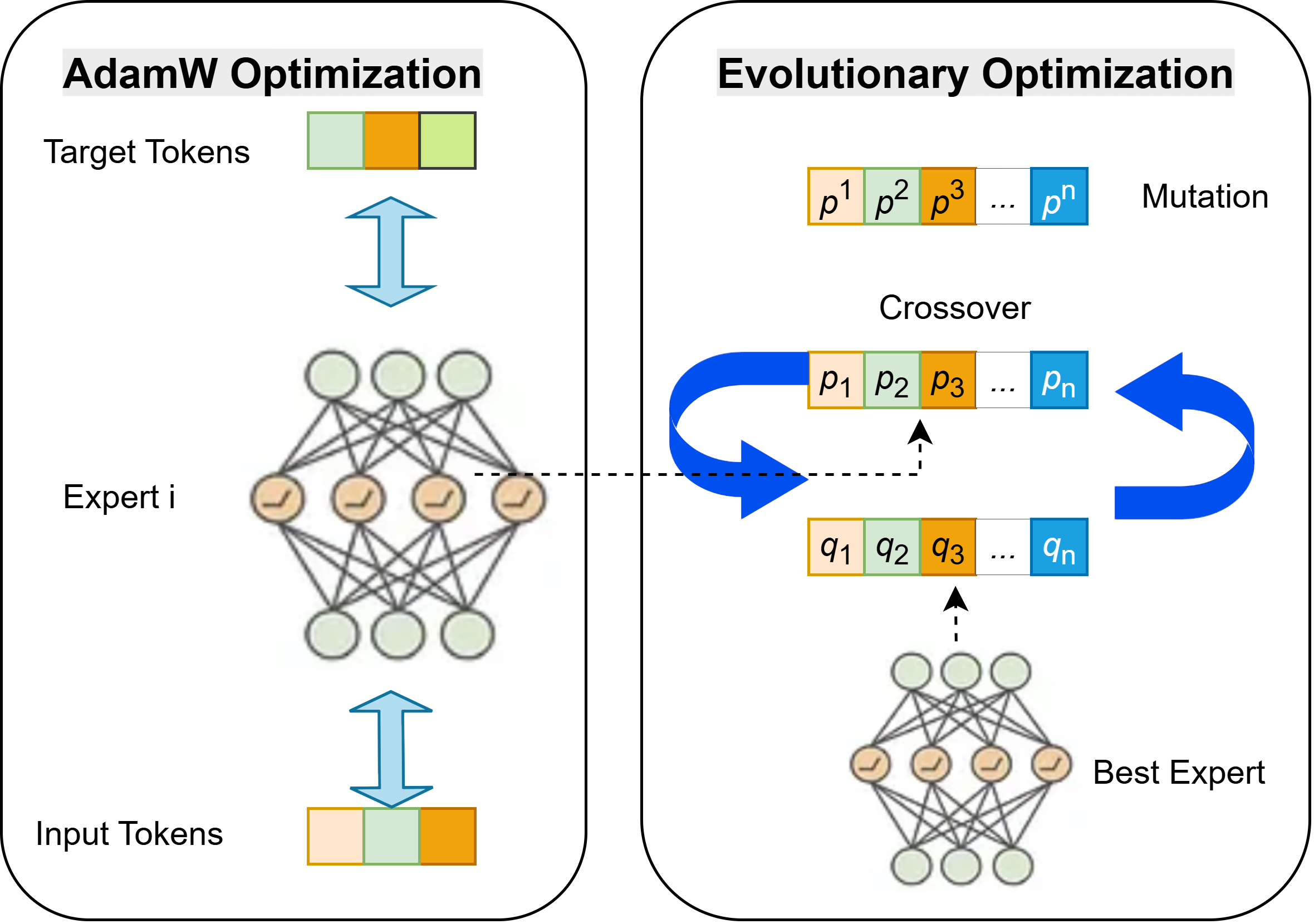}
    \caption{An evolutionary optimization framework for the training of large language models(LLM)}
    \label{fig:eoe}
\end{figure}

This paper presents a new sparse expert framework of evolutionary optimization of experts(EOE). As Figure \ref{fig:eoe} shows, it splits the full model into some equal-sized sub-networks(experts) and trains each sub-network one by one. These experts have the same structure, but their parameter values are different. They are fed with different tokens. They have different search paths and traversed different mountains and valleys in the solution space. Just like different particles of particle optimization algorithm(PSO)\cite{bonyadi2017particle}, or different genes of genetic algorithm(GA)\cite{mitchell1998introduction}, or different ants of ant colony optimization(ACO)\cite{blum2005ant}. So it's natural to follow the evolutionary optimization process of these algorithms. Although the evolutionary operators(mutation, crossover, ...) are different from the classical AdamW\cite{loshchilov2017decoupled} method, their essence is similar, that is, to find the model parameters (weights) $\theta$ that minimize the loss function $\mathcal{L}(\theta)$.

Table \ref{tab:moe_eoe} lists some key difference between MoE and EOE framework.
\begin{table} [H]
    \centering
    \caption{Comparison of MoE models and EOE models}
    \label{tab:moe_eoe}
\begin{tabularx}{\linewidth}{|X|c|c|X|} 
    \hline
    \textbf{Model} & \textbf{Division method} & \textbf{Gating network} &\textbf{Inference model} \\
    \hline
     MoE(Mixture of Experts) & Split each FFN to some sub-networks & Yes & Save all parameters \\
    \hline
    EOE(Evolutionary Optimization of Experts) & Split all layers to some sub-networks & No & Only save parameters of best experts \\
    \hline
\end{tabularx}
\end{table}

We have implemented EOE framework in an open source project \textbf{Koifish} at \url{https://github.com/gruai/koifish} \cite{koifish_repo}. Koifish is a pure c++/cu project, which needs much less training resources than other frameworks. The current version could train GPT-2 1558M model on a single 4090 GPU within 37 hours. While in \cite{gpt2_example}, it takes eight H100 GPUs and 24 hours to train a GPT-2 1558M model.

\section{Method}
Algorithm \ref{alg:train_llm} describes the framework of evolutionary optimization of experts. There are mainly three steps: 
\begin{itemize}
    \item Divide model into multiple experts $\experts=\{\experts_1, \experts_2, ..., \experts_N\}$,
    \item Train the current expert using the classic AdamW optimization process,
    \item The current expert learns the experience of the best expert $\experts_{best}$ using the evolutionary optimization process.
\end{itemize}

\begin{algorithm}[t]
\caption{Evolutionary training of LLM with multiple experts}\label{alg:train_llm}
\begin{algorithmic}[1]
\Require Training dataset $\dataset = \{\mathbf{x}_1, \mathbf{x}_2, ...\}$. A batch $\batch$ includes some shuffled token sequence $\mathbf{x}_i$.
\Require A model divided into multiple experts $\experts=\{\experts_1, \experts_2, ..., \experts_N\}$ and $\experts_{best}$ is the best expert with minimal loss.
\Require Hyperparameter: learning rate $\eta$, weight decay $\lambda$, $\epsilon$, crossover ratio $r_c$, mutation ratio $r_m$,  ...
\Ensure Trained model parameters $\params^*$.

\For{each expert $ \experts_i $ in $\experts$}
    \State {Load the parameters of $ \experts_i $}
    \State {Train this expert with some batches $\{\batch_1, \batch_2, ..., \}$}
    
    \For {each batch $\batch_k$} 
        \State \textbf{AdamW Optimization:}
        \State Compute loss $\mathcal{L}$ on $\batch_k$     \Comment{Forward pass }
        \State $\texttt{grads} \gets \nabla_{\params} \mathcal{L}$ \Comment{backpropagation}
        \State $\theta_{\experts_i}^{(k+1)} \gets \theta_{\experts_i}^{(k)} - \eta \cdot \widehat{\mathbf{m}}_{\experts_i} / (\sqrt{\widehat{\mathbf{v}}_{\experts_i}} + \epsilon) - \eta \lambda \theta_{\experts_i}^{(k)}$
        
        \State \textbf{Evolutionary Optimization:}
        \State $\texttt{grads} \gets \text{Mutation}(\texttt{grads}, {r_m})$
        \State $\texttt{grads} \gets \text{CrossOver}(\texttt{grads of $\experts_{best}$} , {r_c})$
        \State \textbf{Update best expert:}
        \If{$\mathcal{L}<\mathcal{L}_{best}$}    
            \State $\experts_{best}$ = $ \experts_i $
        \EndIf

    \EndFor
\EndFor

\State \Return $\params$
\end{algorithmic}
\end{algorithm}

\subsection{Divide model into multiple experts}

 A decoder-only transformer\cite{vaswani2017attention,chen2021brownian} model $\mathcal{N}$ consists three component:
 \begin{enumerate}
 \item Input component: A token embeddings layer( may followed by positional encoding as GPT-2).  Its parameters are represented by $\theta_{in}$.
 \item Stacked encoder component with $N$ blocks. Each block contains a multi-head self-attention(MHSA) layer and a position-wise Feed-Forward Network(FFN) layer. Koifish splits $N$ block into $K$ partitions. Its parameters $\theta_{encoder}$ are split into $K$ partitions:
 \begin{equation*}
        \theta_{encoder}=\{\theta_{1},\theta_{2}, ..., \theta_{K}\}
 \end{equation*} 
  \item Output component:
  A linear mapping layer(maps the last decoder's output to unnormalized pre-logits) followed by a softmax layer to produce class probabilities. Its parameters are represented by $\theta_{out}$.
  \end{enumerate}

Koifish defines each expert $\experts_i$ as a union of input component, output component, and one partition of encoder component. Its parameters are:  
\begin{equation*}
\theta_{\experts_i}=\{\theta_{in}, \theta_{i}, \theta_{out}\}
 \end{equation*} 
These experts share the same input and output components. They are trained  one by one. For each batch $\batch$, only train and update the parameter of one expert. And the parameters of all other experts are fixed. 

For distinction, we refer to the original model as \textbf{full model} and $\experts=\{\experts_1, \experts_2, ..., \experts_N\}$ as \textbf{EOE model}. 

\subsection{AdamW Optimization}
\begin{enumerate}
    \item \textbf{Forward Pass:} Process a batch $\batch$ to obtain the probability distribution of target tokens.
    \item \textbf{Loss Calculation:} Compute the average cross-entropy loss between the predicted logits and the true target tokens.
        \begin{equation*}
        \mathcal{L} = -\frac{1}{T} \sum_{t=1}^{T} \log P(w_t | w_{<t})
        \end{equation*}
        where $T$ is the batch size, and $w_t$ is the true token at position $t$.

    \item \textbf{Backward Propagation:} Calculate the gradients of the loss. Calculate the gradients of the parameters of current expert $ \experts_i $ via backpropagation.
    \item \textbf{Parameter Update:} Update current expert parameters $\theta_{\experts_i}$ using the computed gradients. At the $k$-th step, a typically used update rule of AdamW\cite{loshchilov2017decoupled} optimizer is:
        \begin{equation*}
        \theta_{\experts_i}^{(k+1)} = \theta_{\experts_i}^{(k)} - \eta \cdot \widehat{\mathbf{m}}_{\experts_i} / (\sqrt{\widehat{\mathbf{v}}_{\experts_i}} + \epsilon) - \eta \lambda \theta_{\experts_i}^{(k)}
        \end{equation*}
where $\widehat{\mathbf{m}}$ and $\widehat{\mathbf{v}}$ are bias-corrected estimates of the first and second moments of the gradients, respectively.
The parameters of all other experts are fixed. 
\end{enumerate}

\subsection{Evolutionary Optimization}
The evolutionary optimization of EOE framework mainly used the following three evolutionary operators:

\begin{enumerate}
\item{\textbf{Particle swarm}}

In many particle swarm algorithms(PSO)\cite{bonyadi2017particle}: individuals always follow the direction of the leaders(best expert). Koifish uses a similar method to update the elements of current expert:
\begin{equation*}
\theta_{\experts_i}^j = \theta_{\experts_i}^j+r_{social} * p * \cdot (\theta_{\experts_{best}}^j -  \theta_{\experts_{i}}^j)
\end{equation*}
where $r_{social}$ is the social component of PSO and $p$ is a random variable between $(0,1)$.

\item{\textbf{Crossover}}

The standard crossover operator of GA\cite{mitchell1998introduction} combines two tensors to produce new offsprings. Koifish adopt a one-way crossover: only current expert would copy some elements from the best expert with a low probability $p$:
\begin{equation*}
\theta_{\experts_i}^j = \theta_{\experts_{best}}^j \;\text{if}\; p<r_c \; \text{otherwise} \; \theta_{\experts_i}^j 
\end{equation*}
where $r_c$ is the ratio of crossover and $p$ is a random variable.

\item{\textbf{Mutation}}

Mutation operator is vital for maintaining diversity in genetic algorithm(GA)\cite{mitchell1998introduction}. More diversity would reduce the possibility of getting stuck in the local minima. GA randomly modifies one or more genes of a chromosome with the mutation probability $p$. Koifish use similar method. Each element of $\theta_{\experts_i}$ has a certain(very small) probability of gaussian mutation:
\begin{equation*}
\theta_{\experts_i}^j = \theta_{\experts_i}^j + \mathcal{N}(0, s) \;\text{if}\; p<r_m \; \text{otherwise} \; \theta_{\experts_i}^j 
\end{equation*}
where 
\begin{itemize}
    \item $\mathcal{N}(0, s)$ is a Gaussian (normal) distributed random variable with variance $s$,      
    \item $p$ is a random variable,   
    \item $r_m$ is the ratio of mutation.
\end{itemize}
A reasonable value of variance $s$ is the second moment $\widehat{\mathbf{v}}$ of AdamW optimization algorithm. Maybe need some scaling factor, which is a hyperparameter to be determined through some experiments.

\end{enumerate}

\section{Experiments}
We have done some experiments on the basis of three GPT-2\cite{radford2019language} models: Small (124M parameters, 12 layers),  Large (774M parameters, 36 layers), and XL (1558M parameters, 48 layers). As Algorithm \ref{alg:train_llm}, Koifish would divide these \textbf{full models} into some \textbf{EOE models}. As shown in Table \ref{tab:eoe_parameter}, each expert has 6 or 8 layers. For example, in the case of GPT-2 XL, EOE model has 6 experts, each has only 328.6M parameters. Koifish would train these experts one by one. At each step, only load and update 328.6M parameters. After some steps, all 1558M parameters are updated. After the training process, Koifish only saves the best model, which only contains 328.6M parameters. It's only one-fifth of the full model file. 

\begin{threeparttable}[ht]
    \centering
    \caption{Parameters of full models and EOE models}
    \label{tab:eoe_parameter}
    \begin{tabular}{|l|c|c|c|c|}
    \hline
        \textbf{Model} & \textbf{Full Parameters} & \textbf{Experts Number} & \textbf{Layers in each expert} & \textbf{Parameters of each expert} \\
    \hline
         GPT2-small & 124M  & 2 & 6 & 81M \\
         GPT2-large & 774M  & 6 & 6 & 183M \\
         GPT2-XL & 1558M  & 6 & 8 & 328.6M\\
    \hline
    \end{tabular}    
\end{threeparttable}

All experiments are conducted on a single NVIDIA 4090 GPU with 24G memory \footnote{All detailed logs are available at \url{https://github.com/gruai/koifish/tree/main/cases/gpt2/}}. The training and test datasets come from the FineWeb Edu 100B dataset\url{karpathy/fineweb-edu-100B-gpt2-token-shards}, which is tokenized with GPT-2 tokenizer using the code in llm.c\cite{llmc_repo}.It includes about 1000 files. Each file contains 100M tokens. Our training only used five billion tokens of FineWeb.

Table \ref{tab:gpt2_summary} lists the training summary.
\begin{enumerate}
\item{GPU Memory}

EOE models need much less resource than full models. For example, in the case of GPT-2 XL, train an EOE model needs ~23G memory, which is only one-fifth of that to train a full model.

\item{Loss}

The loss of EOE models is slightly worse than the baseline that is from llm.c \cite{llmc_repo}( A pure c/cu reproduction of GPT2). We can optimistically estimate that this result can be further improverd. The EOE method has just been proposed, and there are still many ways to enhance its performance.

\item{Throughput}

The throughput obtained using EOE method is very high. In the case of GPT2-XL, the throughput > 50 k/second, which is at least fifty times that of training a full model. Actually, its really hard to train a full model in single 4090. It needs ~118G memory, which is much greater than the 24G memory of 4090.  So have to use some tricks(quantization, re-materialization, reduce batch-size to 1...). No matter what method, the throughput is less than 1 k/second. So in this experiment,  the EOE framework accelerates the throughput more than fifty times! 
\end{enumerate}

\begin{threeparttable}[h]
    \centering      
    \caption{Training summary of three experiments}
    \label{tab:gpt2_summary}  
    \begin{tabular}{|l|c|c|c|c|c|c|}
        \hline
        \textbf{Model} & \textbf{Parameters} & \textbf{Loss (Baseline)} & \textbf{EOE Memory}\tnote{1} & \textbf{Full Memory}\tnote{2} & \textbf{Total time} \tnote{3}  & \textbf{Throughput} \\
        \hline
        GPT2-small & 124M & 3.287 (3.425) & $\sim$6.8G & $\sim$11.9G & $\sim$8 hours & $\sim$140k/s \\
        GPT2-large & 774M & 3.146 (3.00) & $\sim$15G & $\sim$73G & $\sim$18 hours & $\sim$70k/s \\
        GPT2-XL & 1558M & 3.04 (2.83) & $\sim$23G &  $\sim$118G &$\sim$30 hours & $\sim$50k/s \\
        \hline
    \end{tabular}
    \begin{tablenotes}
        \item[1] EOE Memory is the amount of GPU memory needed to train EOE models.
        \item[2] Full Memory is the amount of GPU memory needed to train Full models.
        \item[3] Total time includes both training time and test time. We use only $~10\%$ randomly sampled tokens to reduce test time.

      \end{tablenotes}
    
\end{threeparttable}

\begin{figure}
    \centering
    \includegraphics[width=1.0\linewidth]{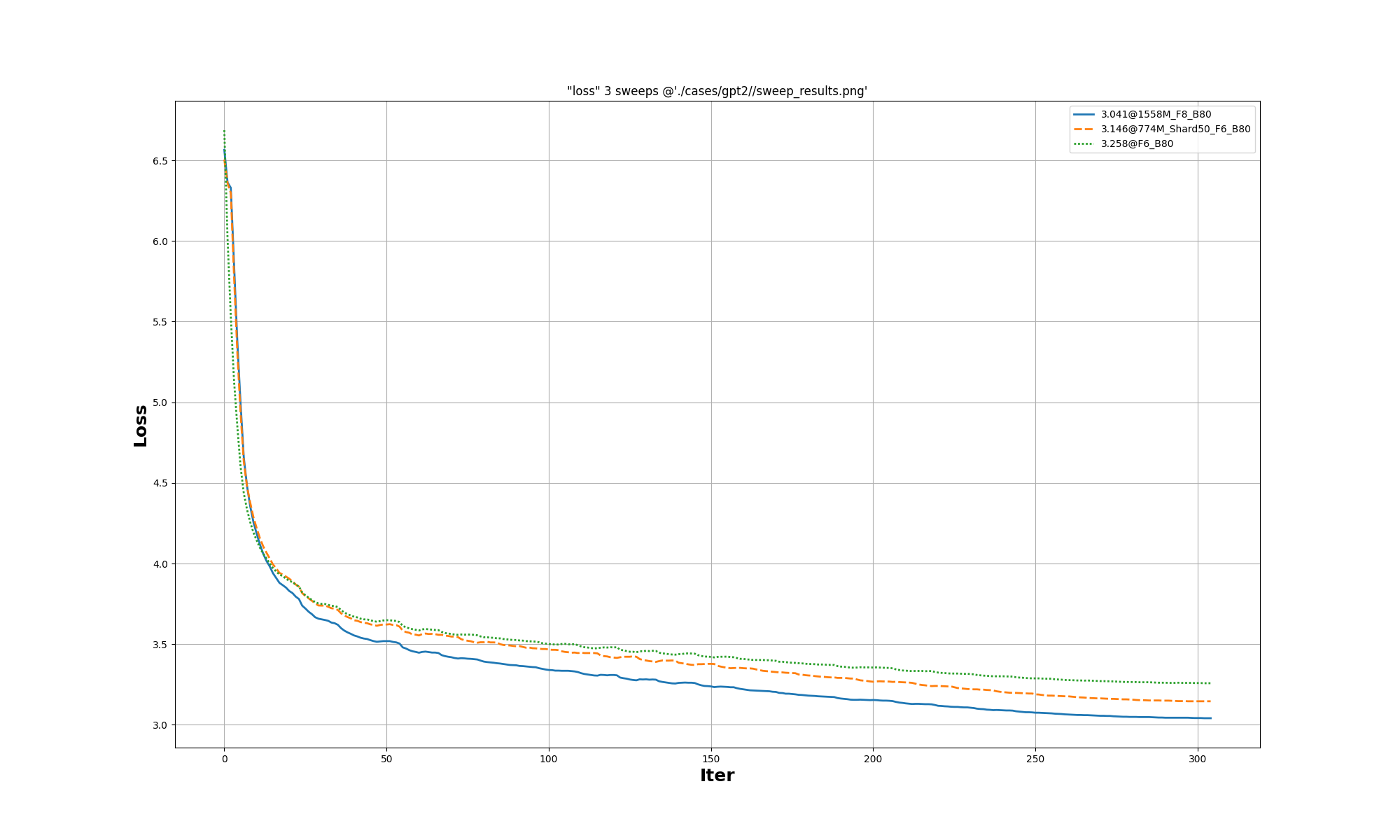}
    \caption{Experimental Results}
    \label{fig:placeholder}
\end{figure}

\section{Concluding Remarks}
We present a new evolutionary optimization framework for training LLM models. We release an open-source project Koifish, which can train GPT2-1558M on a single 4090 GPU. It's just a starting point, we would try bigger models and report more experimental details. 

A long-standing view is that deeper models(with more layers) would achieve higher accuracy. All layers within classical deep models can only be executed sequentially. So deeper models need more memory to store temporary variables and results.  Our EOE framework is actually a sparse, flattened model. For a model with a fixed number of parameters, more experts means wider and flatter models. Is a wider model could achieve the same accuracy as a deeper model? Is this strategy still effective for a bigger model?

\bibliographystyle{unsrt}  
\bibliography{references}  






\end{document}